\documentclass{article}

\usepackage[dvipsnames]{xcolor}
\usepackage{hyperref}
\usepackage{microtype}
\usepackage{graphicx}
\usepackage{booktabs} 
\usepackage{amsmath}
\usepackage{amssymb}
\usepackage{caption}
\usepackage{paralist}
\usepackage{scalefnt}
\usepackage{xspace}
\usepackage{float}
\usepackage[utf8]{inputenc}
\usepackage{color}
\usepackage{tcolorbox}
\usepackage{titling}
\usepackage{authblk}
\usepackage{microtype}
\usepackage{graphicx}
\usepackage{subfigure}
\usepackage{booktabs} 
\usepackage{natbib}
\definecolor{myblue}{RGB}{23,101,139}
\hypersetup{colorlinks=true,citecolor=MidnightBlue,urlcolor=MidnightBlue,linkcolor=MidnightBlue}

\definecolor{myred}{rgb}{0.8,0,0}
\definecolor{mygreen}{rgb}{0,0.6,0}
\definecolor{myblue}{rgb}{0,0,0.7}

\newcounter{cmes} 
\newcommand{\cmes}{\arabic{cmes}}
\newenvironment{mymes}[1]{\vspace{0.3cm}\begin{tcolorbox}[colback=red!10!white]\refstepcounter{cmes}{\bf Message \cmes:}\label{#1}}{\end{tcolorbox}\vspace{0.3cm}}

\newenvironment{myex}[1]{\vspace{0.3cm} \begin{tcolorbox}[colback=blue!10!white]{\bf {#1}}}{\end{tcolorbox}\vspace{0.3cm}}

\newenvironment{myrecom}[1]{\vspace{0.3cm} \begin{tcolorbox}[colback=red!10!white]{\bf {#1}}}{\end{tcolorbox}\vspace{0.3cm}}
\begin{document}

\title{\vspace{-2cm}How Many Random Seeds? Statistical Power Analysis in Deep Reinforcement Learning Experiments}

\author[1]{C\'edric Colas}
\author[1,2]{Olivier Sigaud}
\author[1]{Pierre-Yves Oudeyer}
\affil[1]{INRIA, Flowers team, Bordeaux, France}
\affil[2]{Sorbonne Universit\'e, ISIR, Paris, France}
\renewcommand\Authands{ and }

\posttitle{\par\end{center}}
\date{\vspace{-5ex}}
\maketitle

\begin{abstract}
Consistently checking the statistical significance of experimental results is one of the mandatory methodological steps to address the so-called ``reproducibility crisis" in deep reinforcement learning. In this tutorial paper, we explain how the number of random seeds relates to the probabilities of statistical errors. For both the t-test and the bootstrap confidence interval test, we recall theoretical guidelines to determine the number of random seeds one should use to provide a statistically significant comparison of the performance of two algorithms. Finally, we discuss the influence of deviations from the assumptions usually made by statistical tests. We show that they can lead to inaccurate evaluations of statistical errors and provide guidelines to counter these negative effects. We make our code available 	to perform the tests\footnote{Available on github at \url{https://github.com/flowersteam/rl-difference-testing}}.
\end{abstract}

\section{Introduction}
Reproducibility in Machine Learning and Deep Reinforcement Learning (RL) in particular has become a serious issue in the recent years. As pointed out in \citep{islam2017reproducibility} and \citep{henderson2017deep}, reproducing the results of an RL paper can turn out to be much more complicated than expected. Indeed, codebases are not always released and scientific papers often omit parts of the implementation tricks. Recently, Henderson et al. conducted a thorough investigation of various parameters causing this reproducibility crisis. They used trendy deep RL algorithms such as DDPG \citep{lillicrap2015continuous}, ACKTR \citep{wu2017scalable}, TRPO \citep{schulman2015trust} and PPO \citep{schulman2017proximal} with OpenAI Gym \citep{brockman2016openai} popular benchmarks such as Half-Cheetah, Hopper and Swimmer, to study the effects of the codebase, the size of the networks, the activation function, the reward scaling or the random seeds. Among other results, they showed that different implementations of the same algorithm with the same set of hyper-parameters led to drastically different results. 

\paragraph{}
Perhaps the most surprising thing is this: running the same algorithm 10 times with the same hyper-parameters using 10 different random seeds and averaging performance over two splits of 5 seeds can lead to learning curves seemingly coming from different statistical distributions. Notably, all the deep RL papers reviewed by \citeauthor{henderson2017deep}. (theirs included) used 5 seeds or less. Even worse, some papers actually report the average of the best performing runs. As demonstrated in \citep{henderson2017deep}, these methodologies can lead to claim that two algorithms performances are different when they are not. A solution to this problem is to use more random seeds, to average more different trials in order to obtain a more robust measure of the algorithm performance. But how can one determine how many random seeds should be used? Shall we use 5, 10 or 100, as in \citep{mania2018simple}? 

\paragraph{}
This work assumes one wants to test a difference in performance between two algorithms. Section~\ref{sec:def} gives definitions and describes the statistical problem of {\em difference testing} while Section~\ref{sec:test} proposes two statistical tests to answer this problem. In Section~\ref{sec:theory}, we present standard guidelines to choose the sample size so as to meet requirements in the two types of {\em statistical errors}. Finally, we challenge the assumptions made in the previous section and propose guidelines to estimate error rates empirically in Section~\ref{sec:assumptions}. The code is available on Github at \url{https://github.com/flowersteam/rl-difference-testing}.

\begin{figure}[H]
  \centering
     {\includegraphics[width=0.92\linewidth]{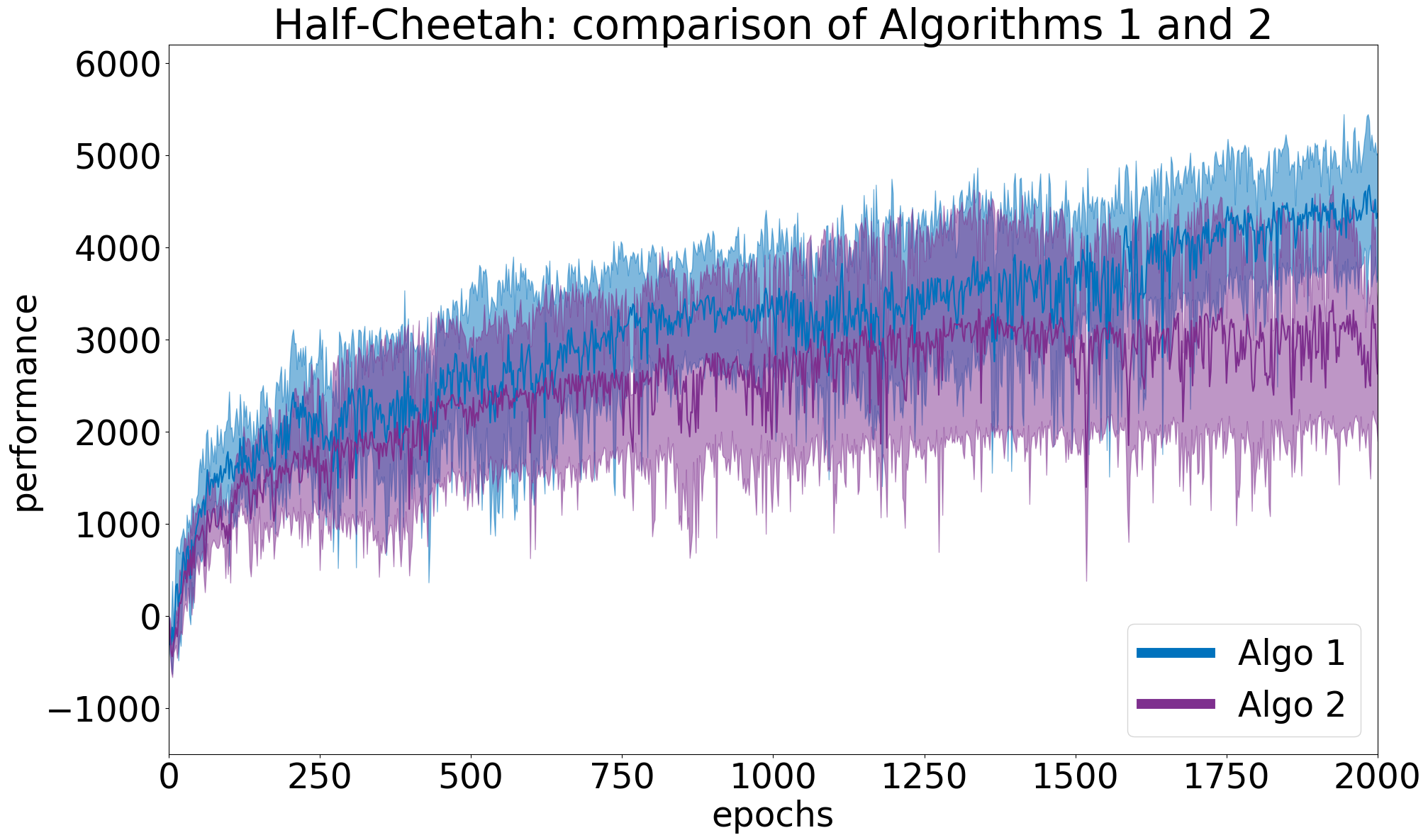} }
    \caption{ \small $Algo1$ versus $Algo2$ are two famous Deep RL algorithms, here tested on the Half-Cheetah benchmark. 
The mean and confidence interval for 5 seeds are reported. We might consider that $Algo1$ outperforms $Algo2$ because there is 
not much overlap between the $95\%$ confidence intervals. But is it sufficient evidence that $Algo1$ really performs better? 
Below, we show that the performances of these algorithms is actually the same, and explain which methods should be used
to have more reliable evidence of the (non-)difference among two algorithms. } \label{fig:ex1_5seeds} 
\end{figure}

\section{Definition of the statistical problem}
\label{sec:def}

\subsection{First definitions}
Two runs of the same algorithm often yield different measures of performance. This might be due to various factors such as the seed of the random generators (called {\em random seed} or {\em seed} thereafter), the initial state of the agent, the stochasticity of the environment, etc.
Formally, the performance of an algorithm can be modeled as a {\em random variable} $X$ and running this algorithm in an environment results in a {\em realization} $x^i$. 
Repeating the procedure $N$ times, one obtains a statistical {\em sample} $x=(x^1, .., x^N)$. A random variable is usually characterized by its {\em expected value} or {\em mean} $\mu$ and its {\em standard deviation}, noted $\sigma$. While the mean characterizes the expected value of a realization, the standard deviation evaluates the square root of the squared deviations to this mean, or in simpler words, how far from the mean the realization are expected to fall. Of course, the values of $\mu$ and $\sigma$ are unknown. The only thing one can do is to compute their unbiased estimations $\overline{x}$ and $s$:

\begin{equation}
\overline{x} \mathrel{\hat=} \sum\limits_{i=1}^n{x^i}, \hspace{1cm} s \mathrel{\hat=}\sqrt{\frac{\sum_{i+1}^{N}(x^i-\overline{x})^2}{N-1}},
\end{equation}
where $\overline{x}$ is called the empirical mean, and $s$ is called the empirical standard deviation. The larger the sample size $N$, the more confidence one can be in the estimations.

\paragraph{}
Here, two algorithms with respective performances $X_1$ and $X_2$ are compared. If $X_1$ and $X_2$ follow normal distributions, the random variable describing their difference $(X_{\textnormal{diff}} = X_1-X_2)$ also follows a normal distribution with parameters ${\sigma_{\textnormal{diff}}=(\sigma_1^2+\sigma_2^2)^{1/2}}$ and $\mu_{\textnormal{diff}}=\mu_1-\mu_2$. In this case, the estimator of the mean of $X_{\textnormal{diff}}$ is $\overline{x}_{\textnormal{diff}} = \overline{x}_1-\overline{x}_2$ and the estimator of ${\sigma_{\textnormal{diff}}}$ is ${s_{\textnormal{diff}}=\sqrt{s_1^2+s_2^2}}$. The {\em effect size} $\epsilon$ can be defined as the difference between the mean performances of both algorithms: ${\epsilon = \mu_1-\mu_2}$. 

\paragraph{}
Testing for a difference between the performances of two algorithms ($\mu_1$ and $\mu_2$) is mathematically equivalent to testing a difference between their difference $\mu_{\textnormal{diff}}$ and 0. The second point of view is considered from now on. We draw a sample $x_{\textnormal{diff}}$ from $X_{\textnormal{diff}}$ by subtracting two samples $x_1$ and $x_2$ obtained from $X_1$ and $X_2$.

\begin{myex}{Example 1.}
To illustrate difference testing, we use two algorithms ($Algo 1$ and $Algo 2$) and compare them on the Half-Cheetah environment from the OpenAI Gym framework \citep{brockman2016openai}. The algorithms implemented are not so important here and will be revealed later. First, we run a study with $N=5$ random seeds for each. Figure~\ref{fig:ex1_5seeds} shows the average learning curves with $95\%$ confidence intervals. Each point of a learning curve is the average cumulated reward over $10$ evaluation episodes. The {\em measure of performance} $X_i$ of $Algo\hspace{3pt}i$ is the average performance over the last $10$ points (i.e. last $100$ evaluation episodes).
From \figurename~\ref{fig:ex1_5seeds}, it seems that $Algo1$ performs better than $Algo2$. Moreover, the confidence intervals do not seem to overlap much. However, we need to run statistical tests before drawing any conclusion.
\end{myex}

\subsection{Comparing performances with a difference test}

In a {\em difference test}, statisticians define the {\em null hypothesis} $H_0$ and the {\em alternate hypothesis} $H_a$. $H_0$ assumes no difference whereas $H_a$ assumes one:

  \begin{itemize}
    \item $H_0$: $\mu_{\textnormal{diff}} = 0$
    \item $H_a$: $\mu_{\textnormal{diff}} \neq 0$ 
  \end{itemize}
These hypotheses refer to the {\em two-tail} case. When an a priori on which algorithm performs best is available, (say $Algo1$), one can use the {\em one-tail} version:

\begin{itemize}
\item $H_0$: $\mu_{\textnormal{diff}} \leq 0$
\item $H_a$: $\mu_{\textnormal{diff}}  > 0$ 
\end{itemize}
At first, a statistical test always assumes the null hypothesis. Once a sample $x_{\textnormal{diff}}$ is collected from $X_{\textnormal{diff}}$, one can estimate the probability $p$ (called $p$-value) of observing data as extreme, under the null hypothesis assumption. By {\em extreme}, one means far from the null hypothesis ($\overline{x}_{\textnormal{diff}}$ far from $0$). The $p$-value answers the following question: {\em how probable is it to observe this sample or a more extreme one, given that there is no true difference in the performances of both algorithms?} Mathematically, we can write it this way for the one-tail case:
\begin{equation}
p{\normalsize \text{-value}} = P(X_{\textnormal{diff}}\geq \overline{x}_{\textnormal{diff}} \hspace{2pt} |\hspace{2pt} H_0),
\end{equation}
and this way for the two-tail case:

\begin{equation}
p{\normalsize \text{-value}}=\left\{
    \begin{array}{ll}
    P(X_{\textnormal{diff}}\geq \overline{x}_{\textnormal{diff}} \hspace{2pt} |\hspace{2pt} H_0)\hspace{0.5cm} \textnormal{if} \hspace{5pt} \overline{x}_{\textnormal{diff}}>0\\
    P(X_{\textnormal{diff}}\leq \overline{x}_{\textnormal{diff}} \hspace{2pt} |\hspace{2pt} H_0) \hspace{0.5cm} \textnormal{if} \hspace{5pt} \overline{x}_{\textnormal{diff}}\leq0.
    \end{array}
    \right.
\end{equation}

When this probability becomes really low, it means that it is highly improbable that two algorithms with no performance difference produced the collected sample $x_{\textnormal{diff}}$. A difference is called {\em significant at significance level $\alpha$} when the $p$-value is lower than $\alpha$ in the one-tail case, and lower than $\alpha/2$ in the two tail case (to account for the two-sided test\footnote{See Wikipedia's article for more details on one-tail versus two-tail tests: \url{https://en.wikipedia.org/wiki/One-_and_two-tailed_tests}}). Usually $\alpha$ is set to $0.05$ or lower. In this case, the low probability to observe the collected sample under hypothesis $H_0$ results in its rejection. Note that a significance level $\alpha=0.05$ still results in $1$ chance out of $20$ to claim a false positive, to claim that there is a true difference when there is not. It is important to note that, when one is conducting $N_E$ experiments, the false positive rate grows linearly with the number of experiments. In this case, one should use correction for multiple comparisons such as the Bonferroni correction $\alpha_{Bon} = \alpha / N_E$ \citep{rice1989analyzing}. This controls the familywise error rate ($FWER$), the probability of rejecting at least one true null hypothesis ($FWER<\alpha$). Its use is discussed in \citep{cabin2000bonferroni}.

\paragraph{}
Another way to see this, is to consider confidence intervals. Two kinds of confidence intervals can be computed:
\begin{itemize}
\item $CI_1$: The $100\cdot(1-\alpha)\hspace{3pt}\%$ confidence interval for the mean of the difference $\mu_{\textnormal{diff}}$ given a sample $x_{\textnormal{diff}}$ characterized by $\overline{x}_{\textnormal{diff}}$ and $s_{\textnormal{diff}}$.
\item $CI_2$: The $100\cdot(1-\alpha)\hspace{3pt}\%$ confidence interval for any realization of $X_{\textnormal{diff}}$ under $H_0$ (assuming $\mu_{\textnormal{diff}}=0$).
\end{itemize}
Having $CI_2$ that does not include $\overline{x}_{\textnormal{diff}}$ is mathematically equivalent to a $p$-value below $\alpha$. In both cases, it means there is less than $100\cdot\alpha\%$ chance that $\mu_{\textnormal{diff}}=0$ under $H_0$. When $CI_1$ does not include $0$, we are also $100\cdot(1-\alpha)\hspace{3pt}\%$ confident that $\mu\neq0$, without assuming $H_0$. Proving one of these things leads to conclude that the difference is {\em significant at level $\alpha$}.

\subsection{Statistical errors}
In hypothesis testing, the statistical test can conclude $H_0$ or $H_a$ while each of them can be either true or false. There are four cases:

\begin{table}[H]
\centering
\caption{Hypothesis testing}
\label{my-label}
\begin{tabular}{c|c|c|}
\cline{2-3}
predicted/true              & $H_0$                                                                & $H_a$                                                                         \\ \hline
\multicolumn{1}{|c|}{$H_0$} & \begin{tabular}[c]{@{}c@{}}True negative\\ $1 - \alpha$\end{tabular} & \begin{tabular}[c]{@{}c@{}}False negative\\ $\beta$\end{tabular} \\ \hline
\multicolumn{1}{|c|}{$H_a$} & \begin{tabular}[c]{@{}c@{}}False positive\\ $\alpha$\end{tabular}    & \begin{tabular}[c]{@{}c@{}}True positive\\ $1-\beta$\end{tabular}             \\ \hline
\end{tabular}
\end{table}

\newpage
\noindent This leads to two types of errors:
\begin{itemize}
\item
The {\bf type-I error} {\bf rejects $H_0$ when it is true}, also called {\em false positive}. This corresponds to claiming the superiority of an algorithm over another when there is no true difference. Note that we call both the significance level and the probability of type-I error $\alpha$ because they both refer to the same concept. Choosing a significance level of $\alpha$ enforces a probability of type-I error $\alpha$, under the assumptions of the statistical test.
\item
The {\bf type-II error} {\bf fails to reject $H_0$ when it is false}, also called {\em false negative}. This corresponds to missing the opportunity to publish an article when there was actually something to be found.
\end{itemize}

\begin{mymes}{mesintro}
\begin{itemize}
\item In the two-tail case, the null hypothesis $H_0$ is $\mu_{\textnormal{diff}}=0$. The alternative hypothesis $H_a$ is $\mu_{\textnormal{diff}}\neq0$.
\item $p{\normalsize \text{-value}} = P(X_{\textnormal{diff}}\geq \overline{x}_{\textnormal{diff}} \hspace{2pt} |\hspace{2pt} H_0)$.
\item A difference is said {\em statistically significant} when a statistical test passed. One can reject the null hypothesis when 1) $p$-value$<\alpha$; 2) $CI_1$ does not contain $0$; 3) $CI_2$ does not contain $\overline{x}_{\textnormal{diff}}$.
\item {\em statistically significant} does not refer to the absolute truth. Two types of error can occur. Type-I error rejects $H_0$ when it is true. Type-II error fails to reject $H_0$ when it is false. 
\item The rate of false positive is 1 out of 20 for $\alpha=0.05$. It grows linearly with the number of experiment $N_E$. Correction procedures can be applied to correct for multiple comparisons.
\end{itemize}
\end{mymes}

\section{Choice of the appropriate statistical test}
\label{sec:test}

In statistics, a difference cannot be proven with $100\%$ confidence. To show evidence for a difference, we use statistical tests. All statistical tests make assumptions that allow them to evaluate either the $p$-value or one of the confidence intervals described in the Section~\ref{sec:def}. The probability of the two error types must be constrained, so that the statistical test produces reliable conclusions. In this section we present two statistical tests for difference testing. As recommended in \cite{henderson2017deep}, the two-sample t-test and the bootstrap confidence interval test can be used for this purpose\footnote{Henderson et al. also advised for the {\bf Kolmogorov-Smirnov test} which tests whether two samples comes from the same distribution. This test should not be used to compare RL algorithms because it is unable to prove any order relation.}.

\subsection{T-test and Welch's t-test}
\label{sec:ttest}

We want to test the hypothesis that two populations have equal means (null hypothesis $H_0$). A 2-sample t-test can be used when the variances of both populations (both algorithms) are assumed equal. However, this assumption rarely holds when comparing two different algorithms (e.g. DDPG vs TRPO). In this case, an adaptation of the 2-sample t-test for unequal variances called Welch's $t$-test should be used \citep{welch1947generalization}. Both tests are strictly equivalent when the standard deviations are equal. $T$-tests make a few assumptions:

\begin{itemize}
\item The scale of data measurements must be continuous and ordinal (can be ranked). This is the case in RL.
\item Data is obtained by collecting a representative sample from the population. This seem reasonable in RL.
\item Measurements are independent from one another. This seems reasonable in RL.
\item Data is normally-distributed, or at least bell-shaped. The normal law being a mathematical concept involving infinity, nothing is ever perfectly normally distributed. Moreover, measurements of algorithm performances might follow multi-modal distributions. In Section,~\ref{sec:assumptions}, we investigate the effects of deviations from normality.
\end{itemize}
Under these assumptions, one can compute the $t$-statistic $t$ and the degree of freedom $\nu$ for the Welch's $t$-test as estimated by the Welch–Satterthwaite equation, such as:

\begin{equation}
t = \frac{x_{\textnormal{diff}}}{\sqrt{\frac{s^2_1}{N_1}+\frac{s^2_2}{N_2}}}, \hspace{1cm} \nu \approx \frac{\Big(\frac{s^2_1}{N_1}+\frac{s^2_2}{N_2}\Big)^2}{\frac{s^4_1}{N^2_1(N_1-1)}+\frac{s^4_2}{N^2_2(N_2-1)}},
\end{equation}
with $x_{\textnormal{diff}} = x_1-x_2$; $s_1, s_2$ the empirical standard deviations of the two samples, and $N_1, N_2$ their sizes. Sample sizes are assumed equal $(N_1=N_2=N)$ thereafter. The $t$-statistics are assumed to follow a $t$-distribution, which is bell-shaped and whose width depends on the degree of freedom. The higher this degree, the thinner the distribution. 

\begin{figure}[H]
  \centering
     {\includegraphics[width=0.9\linewidth]{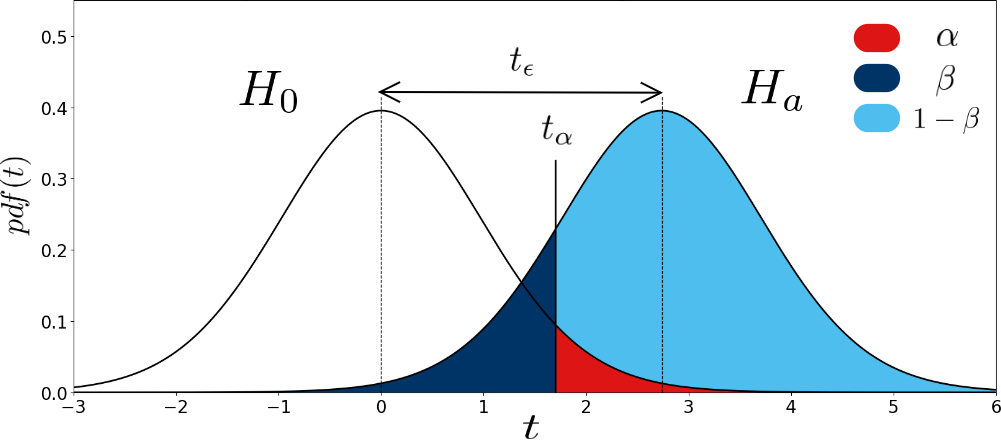} }
    \caption{\small Representation of $H_0$ and $H_a$ under the $t$-test assumptions. Areas under the distributions represented in red, dark blue and light blue correspond to the probability of type-I error $\alpha$, type-II error $\beta$ and the statistical power $1-\beta$ respectively. \label{fig:test_visual}}
\end{figure}

\paragraph{}
\figurename~\ref{fig:test_visual} helps making sense of these concepts. It represents the distribution of the $t$-statistics corresponding to $X_{\textnormal{diff}}$, under $H_0$ (left distribution) and under $H_a$ (right distribution). $H_0$ assumes $\mu_{\textnormal{diff}}=0$, the distribution is therefore centered on 0. $H_a$ assumes a (positive) difference $\mu_{\textnormal{diff}}=\epsilon$, the distribution is therefore shifted by the $t$-value corresponding to $\epsilon$, $t_\epsilon$. Note that we consider the one-tail case here, and test for a positive difference.

\paragraph{}
A $t$-distribution is defined by its {\em probability density function} $T_{distrib}^{\nu}(\tau)$ (left curve in \figurename~\ref{fig:test_visual}), which is parameterized by $\nu$. The {\em cumulative distribution function} $CDF_{H_0}(t)$ is the function evaluating the area under $T_{distrib}^{\nu}(t)$ from $\tau=-\infty$ to $\tau=t$. This allows to write:

\begin{equation}
 p\textnormal{\small-value} = 1-CDF_{H_0}(t) = 1-\int_{-\infty}^{t} T_{distrib}^{\nu}(\tau) \cdot d\tau.
\end{equation}

In \figurename~\ref{fig:test_visual}, $t_\alpha$ represents the critical $t$-value to satisfy the significance level $\alpha$ in the one-tail case. When $t=t_\alpha$, $p$-value$=\alpha$. When $t>t_\alpha$, the $p$-value is lower than $\alpha$ and the test rejects $H_0$. On the other hand, when $t$ is lower than $t_\alpha$, the $p$-value is superior to $\alpha$ and the test fails to reject $H_0$. As can be seen in the figure, setting the threshold at $t_\alpha$ might also cause an error of type-II. The rate of this error ($\beta$) is represented by the dark blue area: under the hypothesis of a true difference $\epsilon$ (under $H_a$, right distribution), we fail to reject $H_0$ when $t$ is inferior to $t_\alpha$. $\beta$ can therefore be computed mathematically using the $CDF$:

\begin{equation}
 \beta = CDF_{H_a}(t_\alpha) = \int_{-\infty}^{t_\alpha} T_{distrib}^{\nu}(\tau-t_{\epsilon}) \cdot d\tau.
\end{equation}
Using the translation properties of integrals, we can rewrite $\beta$ as:

\begin{equation}
 \beta = CDF_{H_0}(t_\alpha-t_{\epsilon}) = \int_{-\infty-t_{\epsilon}=-\infty}^{t_\alpha-t_{\epsilon}} T_{distrib}^{\nu}(\tau) \cdot d\tau.
 \end{equation}

\noindent The procedure to run a Welch's $t$-test given two samples $(x_1, x_2)$ is:

\begin{itemize}
\item Computing the degree of freedom $\nu$ and the $t$-statistic $t$ based on $s_1$, $s_2$, $N$ and $\overline{x}_{\textnormal{diff}}$.
\item Looking up the $t_\alpha$ value for the degree of freedom $\nu$ in a $t$-table\footnote{Available at \url{http://www.sjsu.edu/faculty/gerstman/StatPrimer/t-table.pdf}.} or by evaluating the inverse of the $CDF$ function in $\alpha$. 
\item Compare the $t$-statistic to $t_\alpha$. The difference is said statistically significant ($H_0$ rejected) at level $\alpha$ when $t\geq t_\alpha$.
\end{itemize}

\paragraph{}
Note that $t<t_\alpha$ does not mean there is no difference between the performances of both algorithms. It only means there is not enough evidence to prove its existence with $100 \cdot (1-\alpha)\%$ confidence (it might be a type-II error). Noise might hinder the ability of the test to detect the difference. In this case, increasing the sample size $N$ could help uncover the difference.

\paragraph{}
Selecting the significance level $\alpha$ of the $t$-test enforces the probability of type-I error to $\alpha$. However, \figurename~\ref{fig:test_visual} shows that decreasing this probability boils down to increasing $t_\alpha$, which in turn increases the probability of type-II error $\beta$. One can decrease $\beta$ while keeping $\alpha$ constant by increasing the sample size $N$. This way, the estimation $\overline{x}_{\textnormal{diff}}$ of $\overline{\mu}_{\textnormal{diff}}$ gets more accurate, which translates in thinner distributions in the figure, resulting in a smaller $\beta$. The next section gives standard guidelines to select $N$ so as to meet requirements for both $\alpha$ and $\beta$.

\subsection{Bootstrapped confidence intervals}
Bootstrapped confidence interval is a method that does not make any assumption on the distribution of $X_{\textnormal{diff}}$. It estimates the confidence interval $CI_1$ for $\mu_{\textnormal{diff}}$, given a sample $x_{\textnormal{diff}}$ characterized by its empirical mean $\overline{x}_{\textnormal{diff}}$. It is done by re-sampling inside $x_{\textnormal{diff}}$ and by computing the mean of each newly generated sample. The test makes its decision based on whether the confidence interval of $\overline{x}_{\textnormal{diff}}$ contains $0$ or not. It does not compute a $p$-value as such.

\paragraph{}
Without any assumption on the data distribution, an analytical confidence interval cannot be computed. Here, $X_{\textnormal{diff}}$ follows an unknown distribution $F$. An estimation of the confidence interval $CI_1$ can be computed using the {\em bootstrap principle}.

\paragraph{}
Let us say we have a sample $x_{\textnormal{diff}}$ made of $N$ measures of performance difference. The empirical bootstrap sample $x^*_{\textnormal{diff}}$ of size $N$ is obtained by sampling with replacement inside $x_{\textnormal{diff}}$. The bootstrap principle then says that, for any statistic $u$ computed on the original sample and $u^*$ computed on the bootstrap sample, variations in $u$ are well approximated by variations in $u^*$\footnote{More explanations and justifications can be found in \url{https://ocw.mit.edu/courses/mathematics/18-05-introduction-to-probability-and-statistics-spring-2014/readings/MIT18_05S14_Reading24.pdf}.}. Therefore, variations of the empirical mean such as its range can be approximated by variations of the bootstrapped samples. The bootstrap confidence interval test assumes the sample size is large enough to represent the underlying distribution correctly, although this might be difficult to achieve in practice. Deviations from this assumption are discussed in Section~\ref{sec:assumptions}. Under this assumption, the bootstrap test procedure looks like this: 
\begin{itemize}
\item Generate $B$ bootstrap samples of size $N$ from the original sample $x_1$ of $Algo1$ and $B$ samples from from the original sample $x_2$ of $Algo2$.
\item Compute the empirical mean for each sample: $\mu^1_1, \mu^2_1, ..., \mu^B_1$ and $\mu^1_2, \mu^2_2, ..., \mu^B_2$
\item Compute the differences $\mu_{\textnormal{diff}}^{1:B} = \mu_1^{1:B}-\mu_2^{1:B}$
\item Compute the bootstrapped confidence interval at $100\cdot(1-\alpha)\%$. This is basically the range between the   $100 \cdot\alpha/2$ and $100\cdot(1-\alpha)/2$ percentiles of the vector  $\mu_{\textnormal{diff}}^{1:B}$ (e.g. for $\alpha=0.05$, the range between the $2.5^{th}$ and the $97.5^{th}$ percentiles).
\end{itemize}

\paragraph{}
The number of bootstrap samples $B$ should be chosen large (e.g. $>1000$). If the confidence interval does not contain $0$, it means that one can be confident at $100 \cdot (1-\alpha)\%$ that the difference is either positive (both bounds positive) or negative (both bounds negative), thus, that there is a statistically significant difference between the performances of both algorithms\footnote{An implementation of the bootstrap confidence interval test can be found at \url{https://github.com/facebookincubator/bootstrapped}.}. 

\begin{myex}{Example 1 (continued).}
Here, the type-I error requirement is set to $\alpha=0.05$. Running the Welch's $t$-test and the bootstrap confidence interval test with two samples ($x_1,x_2$) of $5$ seeds each leads to a $p$-value of $0.031$ and a bootstrap confidence interval such that $P\big(\mu_{\textnormal{diff}} \in [259, 1564]\big) = 0.05$. Since the $p$-value is below the significance level $\alpha$ and the $CI_1$ confidence interval does not include $0$, both test passed. This means both tests found a significant difference between the performances of $Algo1$ and $Algo2$ with a $95\%$ confidence. There should have been only $5\%$ chance to conclude a significant difference if it did not exist. 

\paragraph{}
In fact, we did encounter a type-I error. We know this for sure because {\bf {\em Algo 1} and {\em Algo 2} were the exact same algorithm.} They are both the canonical implementation of DDPG \citep{lillicrap2015continuous} from the OpenAI baselines \citep{baselines}. The first conclusion was wrong, we committed a type-I error, rejecting $H_0$ when it was true. We knew this could happen with probability $\alpha=0.05$. Section~\ref{sec:assumptions} shows that this probability might have been under-evaluated because of the assumptions made by the statistical tests.
\end{myex}

\begin{mymes}{mes1}
\begin{itemize}
\item $T$-tests assume $t$-distributions of the $t$-values. Under some assumptions, they can compute analytically the $p$-value and the confidence interval $CI_2$ at level $\alpha$. 
\item The Welch's $t$-test does not assume both algorithms have equal variances but the $t$-test does.
\item The bootstrapped confidence interval test does not make assumptions on the performance distribution and estimates empirically the confidence interval $CI_1$ at level $\alpha$.
\item Selecting a test with a significance level $\alpha$ enforces a type-I error $\alpha$ when the assumptions of the test are verified.
\end{itemize}
\end{mymes}

\section{In theory: power analysis for the choice of the sample size}
\label{sec:theory}

In the Section~\ref{sec:test}, we saw that $\alpha$ was enforced by the choice of the significance level in the test implementation. The second type of error $\beta$ must now be estimated. $\beta$ is the probability to fail to reject $H_0$ when $H_a$ is true. When the effect size $\epsilon$ and the probability of type-I error $\alpha$ are kept constant, $\beta$ is a function of the sample size $N$. Choosing $N$ so as to meet requirements on $\beta$ is called {\em statistical power analysis}. It answers the question: {\em what sample size do I need to have $1-\beta$ chance to detect an effect size $\epsilon$, using a test with significance level $\alpha$?} The next paragraphs present guidelines to choose $N$ in the context of a Welch's $t$-test.

\noindent As we saw in Section \ref{sec:ttest}, $\beta$ can be analytically computed as:

\begin{equation}
\label{eq:beta}
 \beta = CDF_{H_0}(t_\alpha-t_{\epsilon}) = \int_{-\infty-t_{\epsilon}=-\infty}^{t_\alpha-t_{\epsilon}} T_{distrib}^{\nu}(\tau) \cdot d\tau,
\end{equation}
where $CDF_{H_0}$ is the cumulative distribution function of a $t$-distribution centered on $0$, $t_\alpha$ is the critical value for significance level $\alpha$ and $t_\epsilon$ is the $t$-value corresponding to an effect size $\epsilon$. In the end, $\beta$ depends on $\alpha$, $\epsilon$, ($s_1$, $s_2$) the empirical standard deviations computed on two samples ($x_1,x_2$) and the sample size $N$.

\begin{myex}{Example 2.}
To illustrate, we compare two DDPG variants: one with action perturbations ($Algo 1$) \citep{lillicrap2015continuous}, the other with parameter perturbations ($Algo 2$) \citep{plappert2017parameter}. Both algorithms are evaluated in the Half-Cheetah environment from the OpenAI Gym framework \citep{brockman2016openai}.
\end{myex}

\subsection{Step 1 - Running a pilot study}

To compute $\beta$, we need estimates of the standard deviations of the two algorithms ($s_1, s_2$). In this step, the algorithms are run in the environment to gather two samples $x_1$ and $x_2$ of size $n$. From there, we can compute the empirical means $(\overline{x}_1, \overline{x}_2)$ and standard deviations $(s_1, s_2)$.

\begin{figure}[H]
  \centering
     {\includegraphics[width=\linewidth]{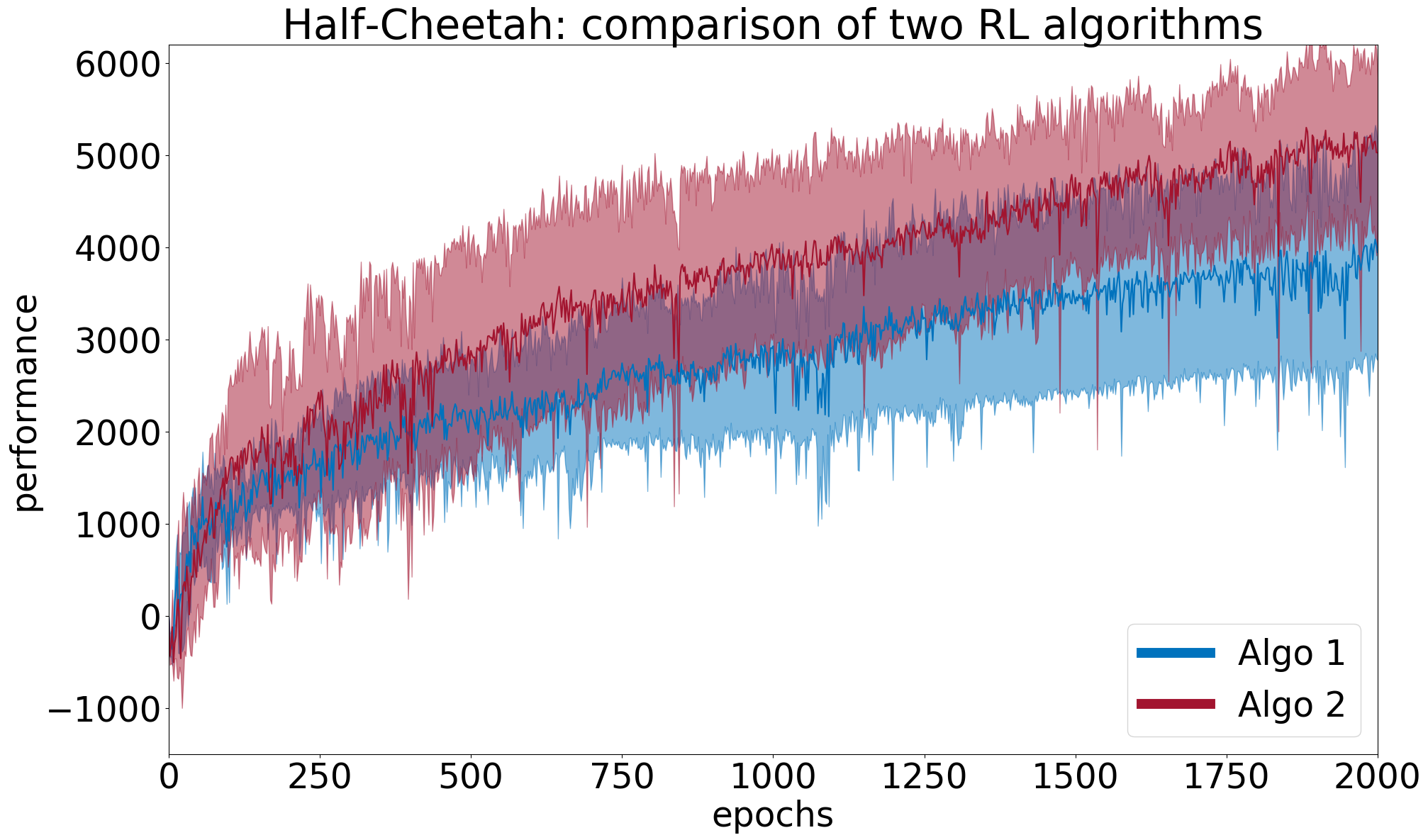} }
    \caption{\small DDPG with action perturbation versus DDPG with parameter perturbation tested in Half-Cheetah. Mean and $95\%$ confidence interval computed over $5$ seeds are reported. The figure shows a small difference in the empirical mean performances. \label{fig:ex2_5}}
  
\end{figure}

\begin{myex}{Example 2 (continued).} 
Here we run both algorithms with $n=5$. We find empirical means $(\overline{x}_1, \overline{x}_2) = (3523, 4905)$ and empirical standard deviations $(s_1, s_2) = (1341, 990)$ for $Algo1$ (blue) and $Algo2$ (red) respectively. From \figurename~\ref{fig:ex2_5}, it seems there is a slight difference in the mean performances  $\overline{x}_{\textnormal{diff}} =\overline{x}_2-\overline{x}_1 >0$.

\paragraph{}
Running preliminary statistical tests at level $\alpha=0.05$ lead to a $p$-value of $0.1$ for the Welch's $t$-test, and a bootstrapped confidence interval of $CI_1=[795, 2692]$ for the  value of $\overline{x}_{\textnormal{diff}} = 1382$. The Welch's $t$-test does not reject $H_0$ ($p$-value$>\alpha$) but the bootstrap test does ($0\not\in CI_1$). One should compute $\beta$ to estimate the chance that the Welch's $t$-test missed an underlying performance difference (type-II error).
\end{myex}

\subsection{Step 2 - Choosing the sample size}
Given a statistical test (Welch's $t$-test), a significance level $\alpha$ (e.g. $\alpha=0.05$) and empirical estimations of the standard deviations of $Algo1$ and $Algo2$ ($s_1,s_2$), one can compute $\beta$ as a function of the sample size $N$ and the effect size $\epsilon$ one wants to be able to detect.

\begin{figure}[H]
  \centering
     {\includegraphics[width=0.8\linewidth]{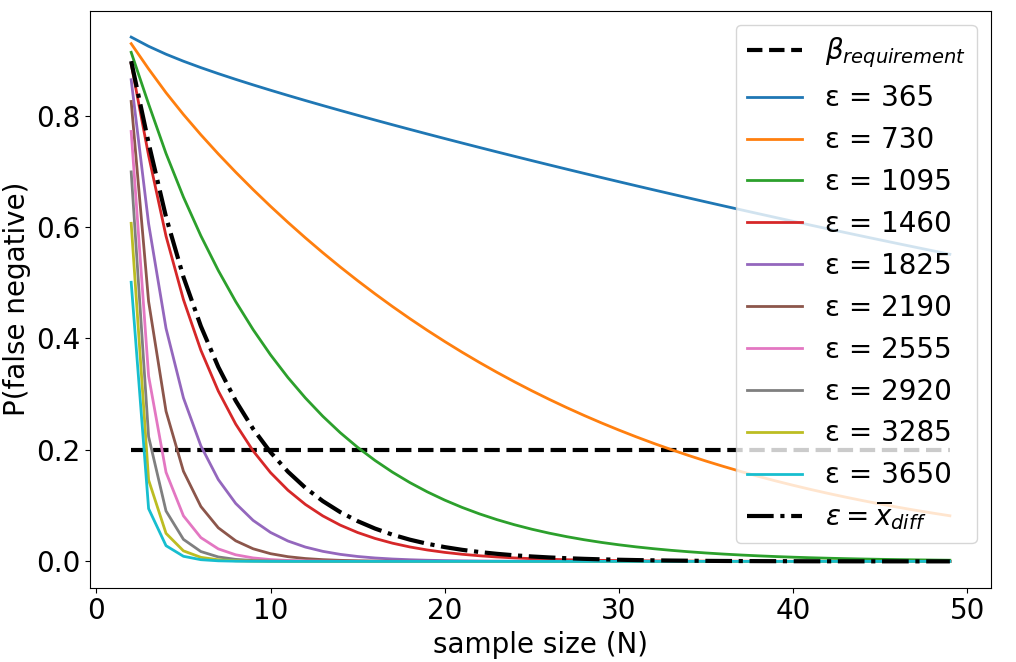} }
    \caption{\small Evolution of the probability of type-II error as a function of the sample size $N$ for various effect sizes $\epsilon$, when $(s_1,s_2)= (1341, 990)$ and $\alpha=0.05$. The requirement $0.2$ is represented by the horizontal dashed black line. The curve for $\epsilon=\overline{x}_{\textnormal{diff}}$ is represented by the semi-dashed black line. \label{fig:beta}}
\end{figure}

\begin{myex}{Example 2 (continued).}
For $N$ in $[2,50]$ and $\epsilon$ in $[0.1,..,1]\times\overline{x}_1$, we compute $t_\alpha$ and $\nu$ using the formulas given in Section \ref{sec:ttest}, as well as $t_{\epsilon}$ for each $\epsilon$. Finally, we compute the corresponding probability of type-II error $\beta$ using Equation~\ref{eq:beta}. \figurename~\ref{fig:beta} shows the evolution of $\beta$ as a function of $N$ for the different $\epsilon$. Considering the semi-dashed black line for $\epsilon=\overline{x}_{\textnormal{diff}}=1382$, we find $\beta=0.51$ for $N=5$: there is $51\%$ chance of making a type-II error when trying to detect an effect $\epsilon=1382$. To meet the requirement $\beta=0.2$, $N$ should be increased to $N=10$ ($\beta=0.19$).
\end{myex}

In our example, we find that $N=10$ was enough to be able to detect an effect size $\epsilon=1382$ with a Welch's $t$-test, using significance level $\alpha$ and using empirical estimations $(s_1, s_2) = (1341, 990)$. However, let us keep in mind that these computations use various approximations ($\nu, s_1, s_2$) and make assumptions about the shape of the $t$-values distribution. Section~\ref{sec:assumptions} investigates the influence of these assumptions.

\subsection{Step 3 - Running the statistical tests}
Both algorithms should be run so as to obtain a sample $x_{\textnormal{diff}}$ of size $N$. The statistical tests can be applied.

\begin{figure}[!ht]
  \centering
     {\includegraphics[width=\linewidth]{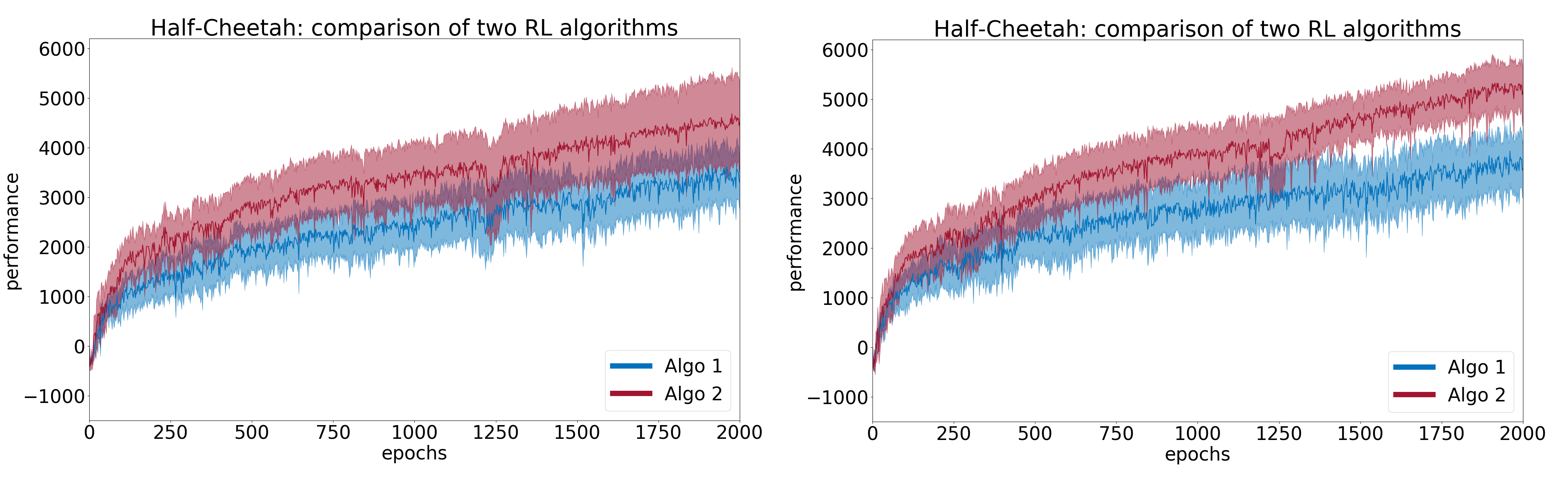} }
    \caption{\small Performance of DDPG with action perturbation ($Algo1$) and parameter perturbation ($Algo2$) with $N=5$ seeds (left) and $N=10$ seeds (right). The $95\%$ confidence intervals on the right are smaller, because more evidence is available ($N$ larger). The underlying difference appears when $N$ grows. \label{fig:ex2}}
\end{figure}

\begin{myex}{Example 2 (continued).}
Here, we take $N=10$ and run both the Welch's $t$-test and the bootstrap test. We now find empirical means $(\overline{x}_1, \overline{x}_2) = (3690, 5323)$ and empirical standard deviations $(s_1, s_2) = (1086, 1454)$ for $Algo1$ and $Algo2$ respectively. Both tests rejected $H_0$, with a $p$-value of $0.0037$ for the Welch's $t$-test and a confidence interval for the difference $\mu_{\textnormal{diff}} \in [732,2612]$ for the bootstrap test. Both tests passed. In \figurename~\ref{fig:ex2}, plots for $N=5$ and $N=10$ can be compared. With a larger number of seeds, the difference that was not found significant with $N=5$ is now more clearly visible. With a larger number of seeds, the estimate $\overline{x}_{\textnormal{diff}}$ is more robust, more evidence is available to support the claim that $Algo2$ outperforms $Algo1$, which translates to tighter confidence intervals represented in the figures.
\end{myex}

\begin{mymes}{last} 
Given a sample size $N$, a minimum effect size to detect $\epsilon$ and a requirement on type-I error $\alpha$ the probability of type-II error $\beta$ can be computed. This computation relies on the assumptions of the $t$-test. 
The sample size $N$ should be chosen so as to meet the requirements on $\beta$. 
\end{mymes}

\section{In practice: influence of deviations from assumptions}
\label{sec:assumptions}
Under their respective assumptions, the $t$-test and bootstrap test enforce the probability of type-I error to the selected significance level $\alpha$. These assumptions should be carefully checked, if one wants to report the probability of errors accurately. First, we propose to compute an empirical evaluation of the type-I error based on experimental data, and show that: 1) the bootstrap test is sensitive to small sample sizes; 2) the $t$-test might slightly under-evaluate the type-I error for non-normal data. Second, we show that inaccuracies in the estimation of the empirical standard deviations $s_1$ and $s_2$ due to low sample size might lead to large errors in the computation of $\beta$, which in turn leads to under-estimate the sample size required for the experiment.

\subsection{Empirical estimation of the type-I error}
Remember, type-I errors occur when the null hypothesis ($H_0$) is rejected in favor of the alternative hypothesis $(H_a)$, $H_0$ being correct. Given the sample size $N$, the probability of type-I error can be estimated as follows:

\begin{itemize}
\item Run twice this number of trials ($2 \times N$) for a given algorithm. This ensures that $H_0$ is true because all measurements come from the same distribution.
\item Get average performance over two randomly drawn splits of size $N$. Consider both splits as samples coming from two different algorithms. 
\item Test for the difference of both fictive algorithms and record the outcome. 
\item Repeat this procedure $T$ times (e.g. $T=1000$)
\item Compute the proportion of time $H_0$ was rejected. This is the empirical evaluation of $\alpha$.
\end{itemize}

\begin{figure}[H]
  \centering
     {\includegraphics[width=0.80\linewidth]{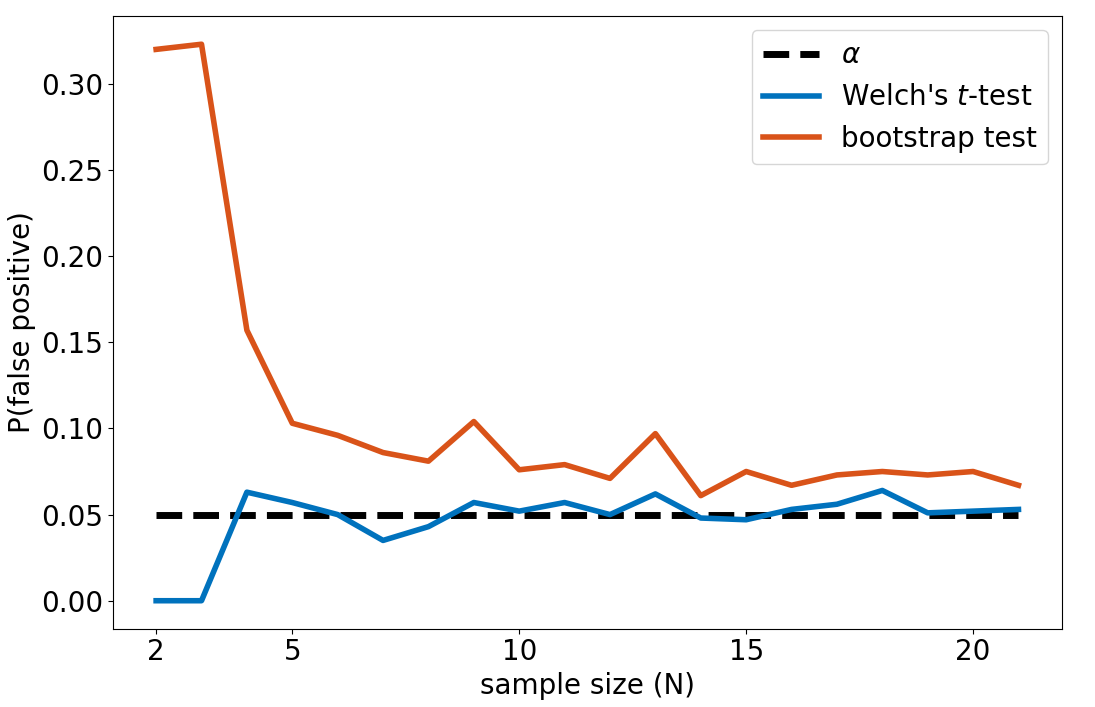} }
    \caption{\small Empirical estimations of the false positive rate on experimental data (Example 3) when $N$ varies, using the Welch's $t$-test (blue) and the bootstrap confidence interval test (orange).  \label{fig:empirical_alpha}}
\end{figure}

\begin{myex}{Example 3}
We use $Algo1$ from Example 2. From $42$ available measures of performance, the above procedure is run for $N$ in $[2,21]$. \figurename~\ref{fig:empirical_alpha} presents the results. For small values of $N$, empirical estimations of the false positive rate are much larger than the supposedly enforced value $\alpha=0.05$.
\end{myex}

\paragraph{}
In our experiment, the bootstrap confidence interval test should not be used with small sample sizes ($<10$). Even in this case, the probability of type-I error ($\approx10\%$) is under-evaluated by the test ($5\%$). The Welch's $t$-test controls for this effect, because the test is much harder to pass when $N$ is small (due to the increase of $t_\alpha$). However, the true (empirical) false positive rate might still be slightly under-evaluated. In this case, we might want to set the significance level to $\alpha<0.05$ to make sure the true positive rate stays below $0.05$. In the bootstrap test, the error is due to the inability of small samples to correctly represent the underlying distribution, which impairs the enforcement of the false positive rate to the significance level $\alpha$. Concerning the Welch's $t$-test, this might be due to the non-normality of our data (whose histogram seems to reveal a bimodal distribution). In Example 1, we used $N=5$ and encountered a type-I error. We can see on the \figurename~\ref{fig:empirical_alpha} that the probability of this to happen was around $10\%$ for the bootstrap test and above $5\%$ for the Welch's $t$-test.

\subsection{Influence of the empirical standard deviations}
\begin{figure}[h]
  \centering
     {\includegraphics[width=\linewidth]{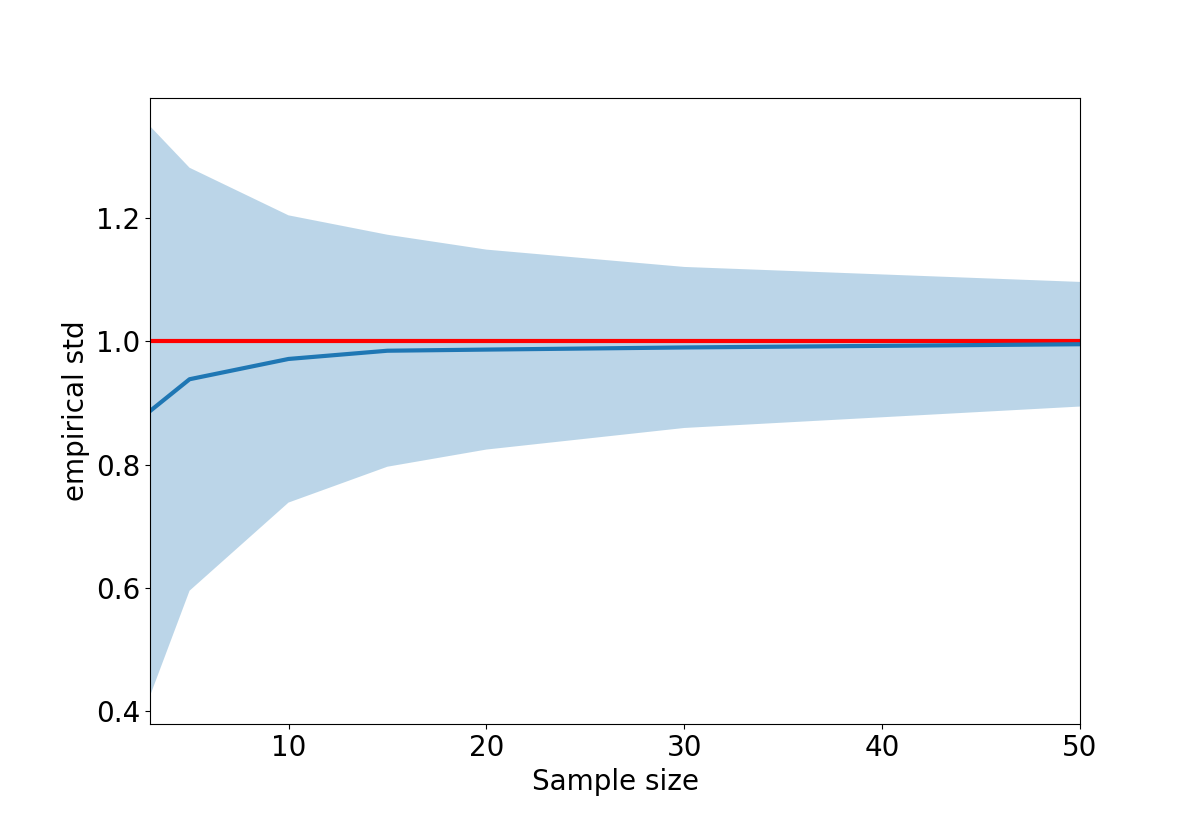} }
    \caption{ \small Empirical standard deviation of $X\sim\mathcal{N}(0,1)$. The true standard deviation $\sigma=1$ is represented in red. Mean +/- std are shown.} \label{fig:std} 
\end{figure}

The Welch's $t$-test computes $t$-statistics and the degree of freedom $\nu$ based on the sample size $N$ and the empirical estimations of standard deviations $s_1$ and $s_2$. When $N$ is low, estimations $s_1$ and $s_2$ under-estimate the true standard deviation in average. Under-estimating $(s_1,s_2)$ leads to smaller $\nu$ and lower $t_\alpha$, which in turn leads to lower estimations of $\beta$. Finally, finding lower $\beta$ leads to the selection of smaller sample size $N$ to meet $\beta$ requirements. Let us investigate how big this effect can be. In \figurename~\ref{fig:std}, one estimates the standard deviation of a normally distributed variable $\mathcal{N}(0,1)$. The empirical estimation $s$ is quite variable and underestimates $\sigma=1$ in average. 

We consider estimations of the false negative rate as a function of $N$ when comparing two normal distributions ($\sigma=1$), one centered on $3$, the other on $3+\epsilon$. When we select $n=5$ for a preliminary study and compute estimations ($s_1,s_2$) from this sample, our average error is mean($s_{n=5}$)$=-0.059$ (see above \figurename~\ref{fig:std}). One could also make larger errors: mean($s_{n=5}$)$-$std($s_{n=5}$)$=-0.40$ from the same figure. 

\begin{figure}[ht]
  \centering
     {\includegraphics[width=\linewidth]{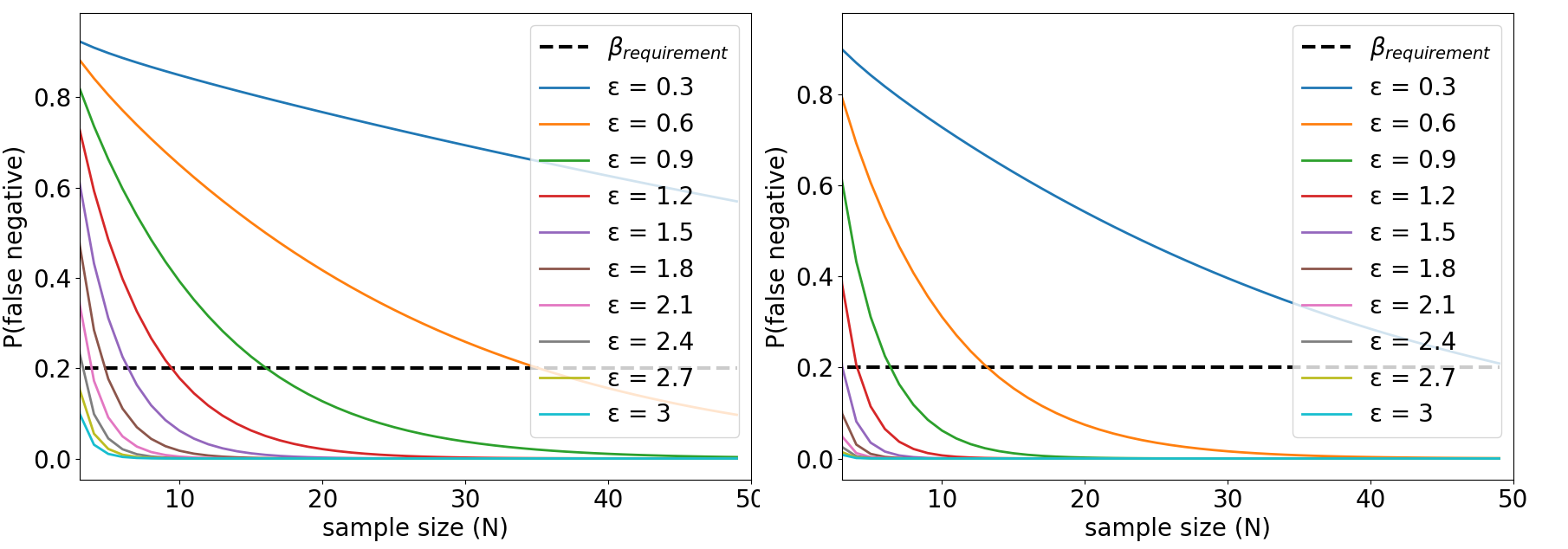} }
    \caption{ \small Evolution of the probability of type-II error as a function of the sample size $N$ and the effect size $\epsilon$, when $(s_1,s_2)= (1-error, 1-error)$ and $\alpha=0.05$. Left: $error=0$, this is the ideal case. Right: $error=0.40$, a large error that can be made when evaluating $s$ over $n=5$ samples. The compared distributions are normal, one is centered on $3$, the other on $3+\epsilon$. \label{fig:combeta} }
\end{figure}

\paragraph{}
\figurename~\ref{fig:combeta} shows the effect of an error of $0.40$ on the evaluation of $\beta$. One can see that, if we want to detect an effect size $\epsilon=0.9$ (green curve) and meet a requirement $\beta=0.2$, one would choose $N=17$ when standard deviations are correctly estimated (left) and $N=7$ when they are under-evaluated. When the number of samples $n$ available in the preliminary study to compute $(s_1,s_2)$ grows, the under-estimation reduces in average and in the worst case. This, in turn, reduces the inaccuracy in the estimation of $\beta$ and therefore in the required $N$. Another solution is to systematically choose the sample size larger than what is prescribed by the computation of $\beta$.

\begin{mymes}{Caution} 
  \begin{itemize}
  	\item One should not blindly believe in statistical tests results. These tests are based on assumptions that are not always reasonable.
    \item $\alpha$ must be empirically estimated, as the statistical tests might underestimate it, because of wrong assumptions about the underlying distributions or because of the small sample size.
    \item The bootstrap test evaluation of type-I error is strongly dependent on the sample size. A bootstrap test should not be used with less than $20$ samples.  
    \item The inaccuracies in the estimation of the standard deviations of the algorithms ($s_1,s_2$), due to small sample sizes $n$ in the preliminary study, lead to under-estimate the sample size $N$ required to meet requirements in type-II errors.
  \end{itemize}
\end{mymes}

\newpage
\section{Conclusion}

In this paper, we outlined the statistical problems that arise when comparing the performance of two RL algorithms. We defined type-I, type-II errors and proposed appropriate statistical tests to test for performance difference. Finally and most importantly, we detailed how to pick the right number of random seeds (the sample size) so as to reach the requirements in both error types. 

\paragraph{}
The most important part is what came after. We challenged the hypotheses made by the Welch's $t$-test and the bootstrap test and found several problems. First, we showed significant difference between empirical estimations of the false positive rate in our experiment and the theoretical values supposedly enforced by both tests. As a result, the bootstrap test should not be used with less than $N=20$ samples and tighter significance level should be used to enforce a reasonable false positive rate ($<0.05$). Second, we show that the estimation of the sample size $N$ required to meet requirements in type-II error were strongly dependent on the accuracy of ($s_1,s_2$). To compensate the under-estimation of $N$, $N$ should be chosen systematically larger than what the power analysis prescribes.

\begin{myrecom}{Final recommendations}
\begin{itemize}
\item Use the Welch's $t$-test over the bootstrap confidence interval test.
\item Set the significance level of a test to lower values ($\alpha<0.05$) so as to make sure the probability of type-I error (empirical $\alpha$) keeps below $0.05$.
\item Correct for multiple comparisons in order to avoid the linear growth of false positive with the number of experiments.
\item Use at least $n=20$ samples in the pilot study to compute robust estimates of the standard deviations of both algorithms. 
\item Use larger sample size $N$ than the one prescribed by the power analysis. This helps compensating for potential inaccuracies in the estimations of the standard deviations of the algorithms and reduces the probability of type-II errors.
\end{itemize}  
\end{myrecom}

\section*{Acknowledgement}
This research is financially supported by the French Minist\`ere des Arm\'ees - Direction G\'en\'erale de l'Armement.

\bibliographystyle{elsarticle-harv}
\end{document}